\documentclass[conference]{IEEEtran}
\usepackage{cite}
\usepackage{amsmath,amssymb,amsfonts}
\usepackage{algorithmic}
\usepackage{graphicx}
\usepackage{subcaption}
\usepackage{textcomp}
\usepackage{xcolor}
\usepackage{standalone}
\usepackage{tikz}
\usepackage{multirow}
\usepackage[vlined, ruled, boxed]{algorithm2e}
\usetikzlibrary{matrix, positioning, calc}
\def\BibTeX{{\rm B\kern-.05em{\sc i\kern-.025em b}\kern-.08em
    T\kern-.1667em\lower.7ex\hbox{E}\kern-.125emX}}
\begin{document}

\title{Bootstrapping a DQN Replay Memory with Synthetic Experiences}

\author{
\IEEEauthorblockN{Wenzel Baron Pilar von Pilchau, Anthony Stein, Jörg Hähner}
\IEEEauthorblockA{
    Organic Computing Group \\
    University of Augsburg, Germany\\
    \textit{\{wenzel.pilar-von-pilchau$~|~$anthony.stein$~|~$joerg.haehner\}}@informatik.uni-augsburg.de
    }
}

\maketitle

\begin{abstract}
An important component of many Deep Reinforcement Learning algorithms is the Experience Replay which serves as a storage mechanism or memory of made experiences. These experiences are used for training and help the agent to stably find the perfect trajectory through the problem space. The classic Experience Replay however makes only use of the experiences it actually made, but the stored samples bear great potential in form of knowledge about the problem that can be extracted. We present an algorithm that creates synthetic experiences in a nondeterministic discrete environment to assist the learner. The Interpolated Experience Replay is evaluated on the FrozenLake environment and we show that it can support the agent to learn faster and even better than the classic version.
\end{abstract}


\section{Introduction}

The concept known as Experience Replay (ER) started as an extension to Q-Learning and AHC-Learning \cite{Lin1992} and developed to a norm in many Deep Reinforcement Learning (RL) algorithms \cite{2015arXiv151105952S,mnih2015human,NIPS2017_7090}. One major advantage is its ability to increase sample efficiency, but algorithms like Deep Q-Network (DQN) are even not able to learn in a stable manner without this extension. This effect is caused by correlations in the observation sequence and the fact that small updates may significantly change the policy and in turn alternate the distribution of the data. By uniformly sampling  over the stored transitions, ER is able to remove these correlations as well as smoothing over changes in the data distribution \cite{mnih2015human}. 

Most versions of ER store the real, actually made, experiences. For instance the authors of \cite{mnih2015human} used vanilla ER for their combination with DQN, and also \cite{2015arXiv151105952S} who extended vanilla ER to their Prioritized Experience Replay, which is able to favor experiences from which the learner can benefit most. But there are also approaches which are filling their replay memory with some kind of synthetic experiences to support the learning process.
One example is the Hindsight Experience Replay from \cite{NIPS2017_7090} which takes a trajectory of states and actions aligned with a goal and replaces the goal with the last state of the trajectory to create a synthetic experience. Both the actual experienced trajectory as well as the synthetic one are then stored in the ER. This approach helps the learner to understand how it is able to reach different goals. This approach was implemented in  a multi-objective problem space and after reaching some \textit{synthetic} goals the agent is able to learn how to reach the intended one.

Our contribution is an algorithm which is targeted to improve (Deep) RL algorithms that make use of an ER, like e.g. DQN, DDPG or classic Q-Learning \cite{DBLP:journals/corr/abs-1712-01275}, in nondeterministic and discrete environments by means of creating synthetic experiences utilizing stored real transitions. We can increase sample efficiency as experiences are further used to generate more and even better transitions. The algorithm therefore computes an average value of the received rewards in a situation and combines this value with observed followup states to create so called \textit{interpolated experiences} which assists the learner in its exploration phase.

The evaluation is performed on the \textit{FrozenLake} environment from the OpenAI Gym \cite{openAI}.

The paper is structured as follows: We start with  a brief introduction of the ER and Deep Q-Learning in \ref{sec:bak} and follow up with some related work in \ref{sec:relw}. In \ref{sec:ier} we introduce our algorithm along with a problem description and the Interpolation Component (IC) which was used as an underlying architecture. The evaluation and corresponding discussion as well as interpretation of the results is presented in \ref{sec:eval} and a conclusion and some ideas of how to proceed further on concludes the paper in \ref{sec:conc}.

\section{Background}
\label{sec:bak}

\subsection{Experience Replay}
\label{sec:er}
ER is a biological inspired mechanism \cite{mcclelland1995there,ONEILL2010220,Lin1992, lin1993reinforcement} to store experiences and reuse them for training later on. An experience therefore is defined as: $e_t=(s_t,a_t,r_r,s_{t+1})$ and stores the state in which the agent started $s_t$, the action it performed $a_t$, the reward it received $r_t$ and the following state it ended up in $s_{t+1}$. At each time step $t$ the agent stores its recent  experience in a replay memory $D_t=\{e_1,\dots,e_t\}$. This procedure is repeated over many episodes, where the end of an episode is defined by a terminal state. The stored transitions can then be utilized for training either online or in a specific training phase. It is very easy to implement ER in its basic form and the cost of using it is mainly determined by the storage space needed.

\subsection{Deep Q-Learning}

The DQN algorithm is the combination of the classic Q-Learning \cite{sutton2018reinforcement} with neural networks and was introduced in \cite{mnih2015human}. The authors showed that their algorithm is able to play Atari 2600 games on a professional human level utilizing the same architecture, algorithm and hyperparameters for every single game.
As DQN is a derivative of classical Q-Learning it approximates the optimal action-value function:
\begin{equation}
    \label{qeq}
    Q^*(s,a)=\max_\pi \mathbb{E}\bigl[r_t+\gamma r_{t+1}+\gamma^2r_{t+2}+\dots|s_t=s,a_t=a,\pi \bigr]
\end{equation}
However DQN employs a neural network instead of a table. Equation \eqref{qeq} displays the maximum sum of rewards $r_t$ discounted by $\gamma$ at each time-step $t$, which is achievable by a behavior policy $\pi=P(a|s)$, after making an observation $s$ and taking an action $a$.
DQN performs an Q-Learning update at every time step which uses the temporal-difference error defined as follows:
\begin{equation}
    \label{eq:td}
    \delta_t = r_t+\gamma \max_{a'}Q(s_{t+1},a')-Q(s_t,a_t) 
\end{equation}

Tsitsiklis and Van Roy \cite{580874} showed that a nonlinear function approximator used in combination with temporal-difference learning, such as Q-Learning, can lead to unstable learning or even divergence of the Q-Function. As a neural network is a nonlinear function approximator, there arise several problems: 1. the correlations present in the sequence of observations, 2. the fact that small updates to Q may significantly change the policy and therefore impact the data distribution, and 3. the correlations between the action-values $Q(s_t,a_t)$ and the target values $r+\gamma\max_{a'}Q(s_{t+1},a')$ present in the td-error shown in \eqref{eq:td}. The last point is crucial, because an update to $Q$ will change the values of both, the action-values as well as the target values, which could lead to oscillations or even divergence of the policy. To counteract these issues, two concrete actions have been proposed: 1. The use of an ER solves, as stated above, the two first points. Training is performed each step on minibatches of experiences $(s,a,r,s')\sim U(D)$, which are drawn uniformly at random from the ER. 2. To remove the correlations between the action-values and the target values a second neural network is introduced which is basically a copy of the network used to predict the action-values, but it is freezed for a certain interval $C$ before it is updated again. This network is called target network and is used for the computation of the target action-values. \cite{mnih2015human}

\section{Related Work}
\label{sec:relw}

The classical ER, introduced in Sec. \ref{sec:er}, is a basic and not optimized technique, which has been improved in many further publications. One prominent improvement is the so called Prioritized Experience Replay \cite{2015arXiv151105952S} which replaces the uniform sampling with a weighted sampling in favor of samples which might influence the learning process most. This modification of the distribution in the replay induces bias and to account for this, importance-sampling has to be used. The authors show that a prioritized sampling leads to great success. Because of the fact that replays store a lot of experiences and the sampling occurs in every training step, it is crucial to reduce the computation cost to a minimum. 

Another publication from de Bruin et al. \cite{de2015importance} investigates the composition of samples in the ER. They discovered that for some tasks it is important, that transitions, made in an early phase when exploration is high, are important to prevent overfitting. Therefore they split the ER in two parts, one with samples from the beginning and one with actual samples. They also show that the composition of the data in an ER is vital for the stability of the learning process and at all times diverse samples should be included.

Jiang et al. investigated ERs combined with model-based RL and implemented a tree structure to represent the transition and reward function \cite{8598780}. In their research they learned a model of the problem and invented a tree structure to represent it. Using this model they could simulate virtual experiences which they used in the planning phase to support learning. To increase sample efficiency, samples are stored in an ER. This approach has some similarities to the interpolation-based approach as presented in this work, but addresses other aspects such as learning a model of the problem first.

An interpolation of on-policy and off-policy model-free Deep Reinforcement Learning techniques present Gu et al. \cite{gu2017interpolated}. In this publication an approach of interpolation between on- and off-policy gradient mixes likelihood ratio gradient with Q-Learning which provides unbiased but high-variance gradient estimations. This approach does not use an ER and therefore differs from our work.

Stein et al. use interpolation in combination with XCS Classifier System to speed up learning in single-step problems by means of using previous experiences as sampling points for the interpolation \cite{Stein2016a,stein2017interpolation,Stein2018}. They introduce a so called Interpolation Component which this work uses as basis for its interpolation tasks.

\section{Interpolated Experience Replay}
\label{sec:ier}

In this section we present the investigated problem and introduce our algorithm to solve it. We also introduce the IC which serves as architectural concept. 

\subsection{Problem description}

In a nondeterministic world an action $a_t\in A$ realised in a state $s_t\in S$ may not lead consistently to the same following state $s_{t+1}\in S$. An example for such an environment would be the ``FrozenLake'' which is basically a grid world consisting of a initial state \textit{I}, a final state \textit{G} and frozen, as well as unfrozen tiles. The unfrozen tiles equal holes \textit{H} in the lake and if the agent falls into one of such, he receives a reward of -1 and has to start from the initial state again. If the agent reaches the final state \textit{G} he receives a reward of 1. The set of possible actions A consists of the four cardinal directions $A=\{N,E,S,W\}$. The environment is nondeterministic, because the agent might slide on the frozen tiles which is implemented through a certain chance of executing a different action instead of the intended one. The environment is discrete, because there is a discrete number of states the agent can reach. The environment used for evaluation is the ``FrozenLake8x8-v0'' environment from OpenAI Gym \cite{openAI} depicted in Figure \ref{fig:frozen_lake}.

\begin{figure}
	\centering
	\includegraphics[width=.5\linewidth]{frozenLake8x8.tex}
	\caption{The FrozenLake8x8-v0 environment from OpenAi Gym \cite{openAI}} 
	\label{fig:frozen_lake}
\end{figure}

If an action is chosen which leads the agent in the direction of the goal, but because of the slippery factor it is falling into a hole, he also receives a negative reward and creates the following experience: $e_t=(s_t,a_t,-1,s_{t+1})$. If this experience is used for a $Q$ update it misleadingly shifts the state-action value away from a positive value. We denote the slippery factor for executing a neighboring action as $c_{\text{slip}}$,  the resulting rewards for executing the two neighboring actions as $r_t^{\text{right}}$ and $r_t^{\text{left}}$ and the reward for executing the intended action as $r_t^{\text{int}}$ and can then define the true expected reward for executing $a_t$ in $s_t$ as follows: 
\begin{equation}
    \label{eq:r-true}
    r_t^{\text{exp}}=\frac{c_{\text{slip}}}{2}\cdot r_t^{\text{right}} + \frac{c_{\text{slip}}}{2}\cdot r_t^{\text{left}} + (1-c_{\text{slip}})\cdot r_t^{\text{int}}
\end{equation}

Following \eqref{eq:r-true} we can define the experience which takes the state-transition function into account and which would not confuse the learner as follows: 
\begin{equation}
    \label{eq:e-true}
    e_t^{\text{exp}}=(s_t,a_t,r_t^{\text{exp}},s_{t+1})
\end{equation}

The learner will converge its state-action value $Q_\pi(s_t,a_t)$ after seeing a lot of experiences to: 
\begin{equation}
    \label{eq:q-true}
    Q_\pi(s_t,a_t)=Q^*(s_t,a_t)=r_t^{\text{exp}}+\gamma\max_{a'}Q^*(s_{t+1},a')
\end{equation}

We define the set of all rewards which belong to the experiences that start in the same state $s_t$ and execute the same action $a_t$ as:
\begin{equation}
    \label{eq:sum-ex}
    R_t:=\big\{r_n \in \{ r | (s, a, r, s') \in D_t \wedge a=a_t \wedge s=s_t \}\big\}
\end{equation}{}

In our work we utilize stored transitions from the replay memory to create synthetic experiences with an averaged reward $r_t^{\text{avg}}$ which is as close as possible to $r_t^{\text{exp}}$. Following \eqref{eq:sum-ex} we can define these \textit{interpolated} experiences as:
\begin{equation}
    \label{eq:r-exp}
    r_t^{\text{avg}} = \frac{\sum_{r \in R_t} r}{|R_t|}
\end{equation}{}
\begin{equation}
    \label{eq:e-avg}
    e_t^{\text{avg}}=(s_t,a_t,r_t^{\text{avg}},s_{t+1})
\end{equation}
with
\begin{equation}
    \label{eq-approx}
    e_t^{\text{avg}}\approx e_t^{\text{exp}}
\end{equation}

The accuracy of this interpolation correlates with the amount of transitions stored in the ER and starting in $s_t$ and executing $a_t$. As a current limitation so far, because we need a legal followup state $s_{t+1}$, it is crucial for the environment to be discrete. Otherwise we had to somehow interpolate or predict this following state or else the state-transition function as well and this could harm the accuracy of the interpolated experience.

\subsection{Algorithm}
\label{sec:alg}

Our algorithm triggers an interpolation after every step the agent takes. A query point $x_q\sim U(S)$ is drawn at random from the state space and all matching experiences:
\begin{equation}
    D_{\text{match}}:=\{e_t\in D_t | s_t=x_q\}
\end{equation} 
for which holds that their starting point $s_t$ is equal to the query point $x_q$ are collected from the ER. 
Then for every action $a\in A$ all experiences that satisfy $a_t=a$ are selected from $D_{\text{match}}$ in:
\begin{equation}
    D_{\text{match}}^a:=\{e_t|e_t \in D_{\text{match}} \wedge a_t=a\}    
\end{equation}
The resulting transitions are used to compute an average reward value $r_t^{\text{avg}}$ and a synthetic sample $e_t^{\text{avg}}$ for every distinct next state:
\begin{equation}
    s_{t+1}\in \{s'|(s_t,a_t,r_t,s')\in D_{\text{match}}^a\}    
\end{equation}
is created. This results in a minimum of 0 and a maximum of 3 synthetic experiences per action and sums up to a maximum of 12 synthetic samples per interpolation depending on the amount of stored transitions in the ER. As with the amount of stored real transitions, which can be seen as the collected experience of the model, the quality of the interpolated experiences may get better, a parameter $c_{\text{start\_interpolation}}$ is introduced, which determines the minimum amount of stored experiences before the first interpolation is executed. The associated pseudocode is depicted in Algorithm~\ref{alg:ier}.

\begin{algorithm}
    \caption{Reward averaging in IER}
    \label{alg:ier}
    Initialize $D$\;
    Initialize $D^{\text{inter}}$\;
    \While{$s$ is not terminal state}{
        Store experience $e$ in $D$\;
        Draw $x$ at random from S\;
        Select all $e_t$ that match $s_t=x$ from $D$\;
        Store results in $D_{\text{match}}$\;
        \ForAll{$a\in A$}{
            Select all $e_t$ that match $a_t=a$ from $D_{\text{match}}$\;
            Store results in $D_{\text{match}}^a$\;
            Compute $r_t^{\text{avg}}$\;
            \ForAll{distinct $s_{t+1}$ in $D_{\text{match}}^a$}{
                Create $e_t^{\text{avg}}=(x,a,r_t^{\text{avg}}, s_{t+1})$\;
                Add $e_t^{\text{avg}}$ to $D^{\text{inter}}$\;
            }
        }
    }
\end{algorithm}

\subsection{Interpolation Component}

\begin{figure}
	\centering
	\includegraphics[width=.5\linewidth]{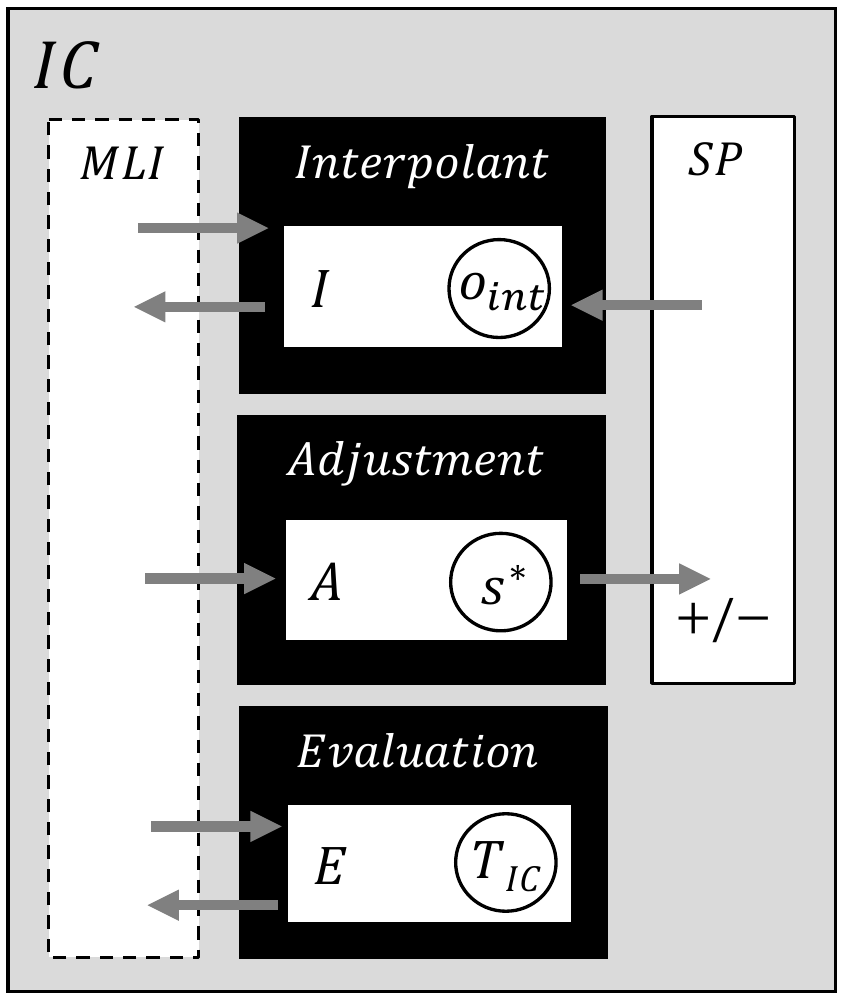}
	\caption{Schematic of the Interpolation Component from Stein et al. \cite{stein2017interpolation}}
	\label{fig:ic}
\end{figure}

Our implementation uses the Interpolation Component from Stein et al. \cite{stein2017interpolation}, depicted  in Fig.~\ref{fig:ic}, as underlying basic structure. The IC serves as an abstract pattern and consists of a Machine Learning Interface (MLI), an Interpolant, an Adjustment Component, an Evaluation Component and the Sampling Points (SP) as shown in Fig.~\ref{fig:ic}. If the MLI receives a sample $s^*$ it is handed to the Adjustment Component, there, following a decision function A, it is added to or removed from SP. If an interpolation is required, the Interpolation Component fetches required sampling points from SP and computes, depending on an interpolation technique I, an output $o_{int}$. The Evaluation Component provides a metric E to track a so-called trust-level $T_{IC}$ as a mertic of interpolation accuracy.

\begin{figure}
	\centering
	\includegraphics[width=.8\linewidth]{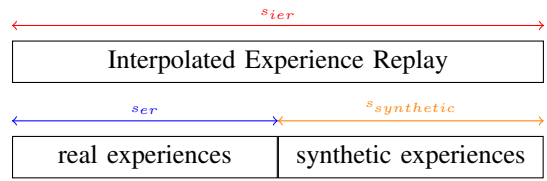}
	\caption{Intuition of Interpolated Experience Replay memory} \label{fig:ier}
\end{figure}

We replaced the SP with the ER. It is realized by a FiFo queue with a maximum length. This queue represents the classic ER and is filled only with real experiences. To store the synthetic samples another queue, a so-called \emph{ShrinkingMemory}, is introduced. This second storage is characterized by a decreasing size. Starting at a predefined maximum it gets smaller depending on the length of the real experience queue. The \emph{Interpolated Experience Replay} (IER) has a total size, comprising the sum of the lengths of both queues as can be seen in Fig.~\ref{fig:ier}. If this size is reached, the length of the ShrinkingMemory is decreased and the oldest items are removed, as long as either the real valued queue reaches its maximum length and there is some space left for interpolated experiences or the IER fills up with real experiences. This approach includes  a minimum size for the interpolated storage, but this was not further investigated in this work and is left for further work.

The IER algorithm as described in \ref{sec:alg} is located in the Interpolant, and, as stated above, executed in every step. The algorithm uses a nearest-neighbor-search with an exact match query to search for all experiences with a matching first state. An exhaustive search would need a computation time of $O(n)$ and therefore is not practical for large sized IERs, because this operation is executed in every single step. A possible solution for this problem is to employ a so called kd-tree, which represents a multidimensional data structure. Using such a tree, the computation time could be decreased to $O(\log N)$ \cite{friedman1977algorithm}. As the examined problem is very small, and consists out of $|S|=64$ discrete states, we use another approach to reduce the computation time further on to $O(1)$. To achieve this we use a dictionary $dict:K\rightarrow V$ of size $|S|*3=192$ with keys:
\begin{equation}
    K:=\{s_t,a_t|s_t\in S, a_t\in A\}
\end{equation}
and corresponding values:
\begin{equation}
    V:=\Big\{r_t^{\text{avg}},\big\{s_{t+1}\in\{s'|(s_t,a_t,r_t,s')\in D_{\text{match}}^a\}|a=a_t\big\}\Big\}    
\end{equation}
This equals an entry for every state-action pair with associated average rewards and distinct next states of all seen transitions. The dictionary is updated after every transition the agent makes. This approach is limited to discrete environments.

To evaluate the quality of computed interpolations in future work, a quality metric could be designed to be used in the Evaluation part of the IC. 

\section{Evaluation}
\label{sec:eval}

\subsection{Experimental setup}

For evaluation purposes, a neural network with only one input and one output and no hidden layer was used. This decision was felt because we use one input node for each state, which gives an overall amount of 64 input nodes. 
Neural networks have the ability to generalize over neighboring areas, but using the architecture described above, this seems to have no effect because every state has its own input node. We therefore decided to reduce complexity by not using hidden layers. 
One output node for every possible action was used, which results in 4 output nodes. Vanilla ER was selected as a baseline and compared with the IER approach presented in Section \ref{sec:ier}. All experiments share the hyperparameters given in Table~\ref{tab:param}. 
Furthermore, different capacities for storing synthetic experiences $s_{syntehtic}$ in combination with different warm-up phases, i.e., values for $c_{start\_interpolation}$, are investigated. 
As exploration technique a linearly decaying $\epsilon$-greedy was used, and different lengths of $t_{exploration}$ tried. 
The different constellations of the individual experiments are shown in Table~\ref{tab:setup}. 
We measure the average reward of the last 100 episodes to obtain a moving average which gives us an impression of how often the agent is able to reach to goal in this time. The problem is considered "solved" when the agent obtains an average reward of at least 0.78 over 100 consecutive episodes.
Each experiment was repeated for 20 times and the results are reported as the mean values and the observed standard deviations ($\pm 1$SD) over the repetitions.
Each configuration was tested against the baseline and the differences have been assessed for statistical significance. Therefore, we first conducted \textit{Shapiro-Wilk} tests in conjuction with visual inspection of \textit{QQ-plots} to determine whether a normal distribution can be assumed. Since this criterion could not be confirmed for any of the experiments, the \textit{Mann-Whitney-U test} has been chosen. All measured statistics, comprising the corresponding p-values for the hypothesis tests are reported in Table~\ref{tab:best}.  

\begin{table}[htpb]
    \centering
	\caption{Overview of hyperparameters applied for the FrozenLake8x8-v0 experiment}
	\begin{tabular}{cc}
		\hline
		\textbf{Parameter} & \textbf{Value} \\
		\hline
		Learning rate $\alpha$ & 0.0005 \\
		Discount factor $\gamma$ & 0.95 \\
		Epsilon start & 1 \\
		Epsilon min & 0.05 \\
		Update target net interval $\tau$ & 300 \\
		Size of Experience Replay $s_{er}$ & 100,000 \\
		Size of Interpolated Experience Replay $s_{ier}$ & 100,000 \\
		Start Learning at size of IER & 300 \\
		Minibatch size & 32 \\
		\hline
	\end{tabular}
	\label{tab:param}
\end{table}

\begin{table}[htpb]
    \centering
    \caption{Overview of the individually conducted experiment constellations}
    \begin{tabular}{c|c|c|c}
         \hline
         \textbf{experiment} & \textbf{$t_{exploration}$} & \textbf{$s_{synthetic}$} & \textbf{$c_{start\_interpolation}$} \\
         \hline
         \multirow{6}{*}{1} & \multirow{6}{*}{500 episodes} & \multirow{3}{*}{20,000} & 250 \\
         & & & 500 \\
         & & & 1,000 \\
         & & \multirow{3}{*}{100,000} & 250 \\
         & & & 500 \\
         & & & 1,000 \\
         \hline
         \multirow{6}{*}{2} & \multirow{6}{*}{750 episodes} & \multirow{3}{*}{20,000} & 250 \\
         & & & 500 \\
         & & & 1,000 \\
         & & \multirow{3}{*}{100,000} & 250 \\
         & & & 500 \\
         & & & 1,000 \\
         \hline
         \multirow{6}{*}{3} & \multirow{6}{*}{1,000 episodes} & \multirow{3}{*}{20,000} & 250 \\
         & & & 500 \\
         & & & 1,000 \\
         & & \multirow{3}{*}{100,000} & 250 \\
         & & & 500 \\
         & & & 1,000 \\
         \hline
    \end{tabular}
    \label{tab:setup}
\end{table}{}

\begin{table*}[htbp]
    \centering
    \caption{Summary of results. Bold entries indicate statistically significant superior performance compared to the baseline.}
    \begin{tabular}{c|c|c|c|c|c|c}
        \hline
        \multirow{2}{*}{\textbf{experiment}} & \multirow{2}{*}{$s_{synthetic}$} & \multirow{2}{*}{$c_{start\_interpolation}$} & \multirow{2}{*}{Mean} & \multirow{2}{*}{$\pm$1SD} & p-value & p-value \\
        & & & & & Shapiro-Wilk & Mann-Whitney-U \\
        \hline
        \multirow{7}{*}{1}  & 0                         & 0     & 0.3772 & $\pm$0.3121 & 1.9085e-33 &  \\
                            \cline{2-7}
                            & \multirow{3}{*}{20,000}   & 250   & \textbf{0.43} & $\pm$0.3277 & 2.0967e-34 & 1.2702e-19 \\
                            &                           & 500   & \textbf{0.4324} & $\pm$0.3297 & 6.5603e-35 & 6.6317e-22 \\
                            &                           & 1,000 & \textbf{0.4416} & $\pm$0.3396 & 2.8203e-34 & 2.4728e-23 \\
                            \cline{2-7}
                            & \multirow{3}{*}{100,000}  & 250   & \textbf{0.4287} & $\pm$0.3364 & 3.611e-35  & 3.2159e-22 \\
                            &                           & 500   & \textbf{0.4168} & $\pm$0.3266 & 9.4551e-35 & 1.4136e-13 \\
                            &                           & 1,000 & \textbf{0.4261} & $\pm$0.3282 & 3.0105e-35 & 9.3601e-21 \\
        \hline
        \multirow{7}{*}{2}  & 0                         & 0     & 0.2385 & $\pm$0.2807 & 8.7954e-35 &  \\
                            \cline{2-7}
                            & \multirow{3}{*}{20,000}   & 250   & \textbf{0.2877} & $\pm$0.3126 & 1.9496e-33 & 1.6066e-06 \\
                            &                           & 500   & \textbf{0.2911} & $\pm$0.3105 & 5.4785e-33 & 1.9508e-06 \\ 
                            &                           & 1,000 & \textbf{0.2653} & $\pm$0.309  & 9.4381e-35 & 1.3255e-02 \\
                            \cline{2-7}
                            & \multirow{3}{*}{100,000}  & 250   & \textbf{0.2785} & $\pm$0.3018 & 5.5895e-33 & 1.7829e-04 \\
                            &                           & 500   & \textbf{0.2782} & $\pm$0.3155 & 1.555e-34  & 2.7468e-03 \\
                            &                           & 1,000 & \textbf{0.2734} & $\pm$0.3026 & 7.4416e-34 & 1.0413e-03 \\
        \hline
        \multirow{7}{*}{3}  & 0                         & 0     & 0.0885 &  $\pm$0.1347 & 5.9146e-38 &  \\
                            \cline{2-7}
                            & \multirow{3}{*}{20,000}   & 250   & \textbf{0.1194} & $\pm$0.1618 & 3.9763e-35 & 3.1267e-09 \\
                            &                           & 500   & \textbf{0.1236} & $\pm$0.1642 & 1.1407e-34 & 6.4606e-09 \\
                            &                           & 1,000 & \textbf{0.1215} & $\pm$0.161  & 8.5940e-35 & 1.0439e-09\\
                            \cline{2-7}
                            & \multirow{3}{*}{100,000}  & 250   & \textbf{0.1198} & $\pm$0.1716 & 1.901e-36  & 2.5456e-03 \\
                            &                           & 500   & \textbf{0.1229} & $\pm$0.1666 & 3.8836e-35 & 3.5305e-03 \\
                            &                           & 1,000 & \textbf{0.116}  & $\pm$0.1602 & 1.5933e-35 & 6.5812e-04\\
        \hline
         
    \end{tabular}
    \label{tab:my_label}
\end{table*}{}

\subsection{Experimental results}

\begin{figure*}
    \centering
    \hspace{-2.5em}
    \begin{subfigure}{.372\linewidth}
        \includegraphics[width=\textwidth]{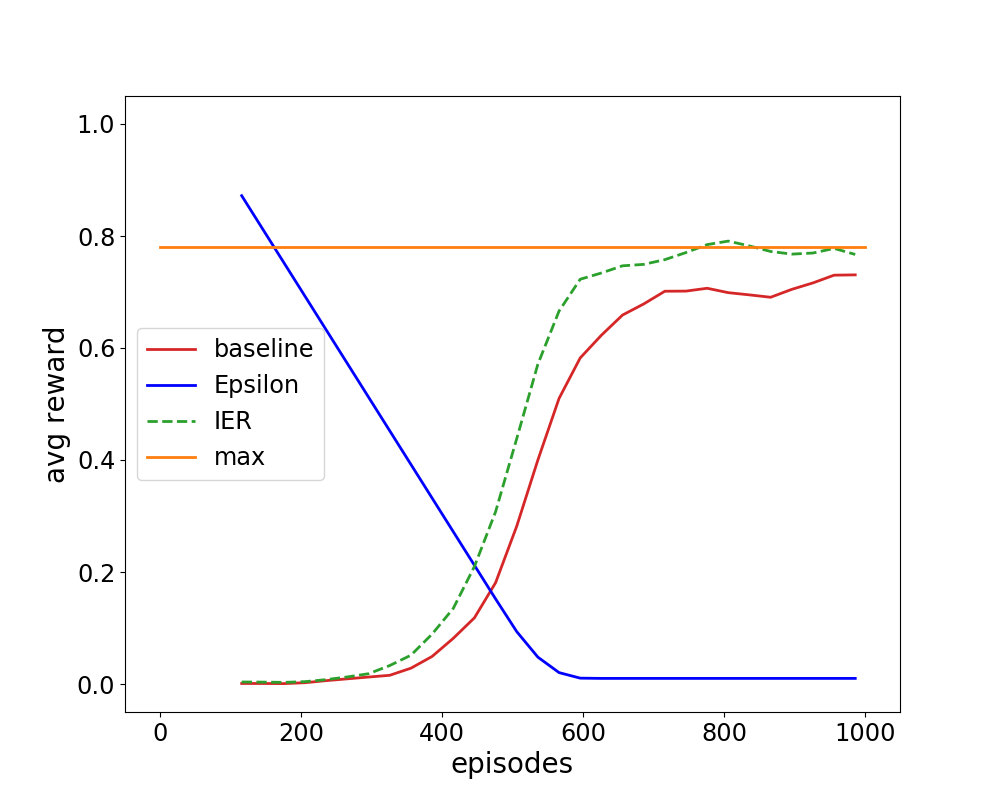}
        \caption{experiment 1}
        \label{fig:exp500}
    \end{subfigure}
    \hspace{-2.34em}
    \begin{subfigure}{.372\linewidth}
        \includegraphics[width=\textwidth]{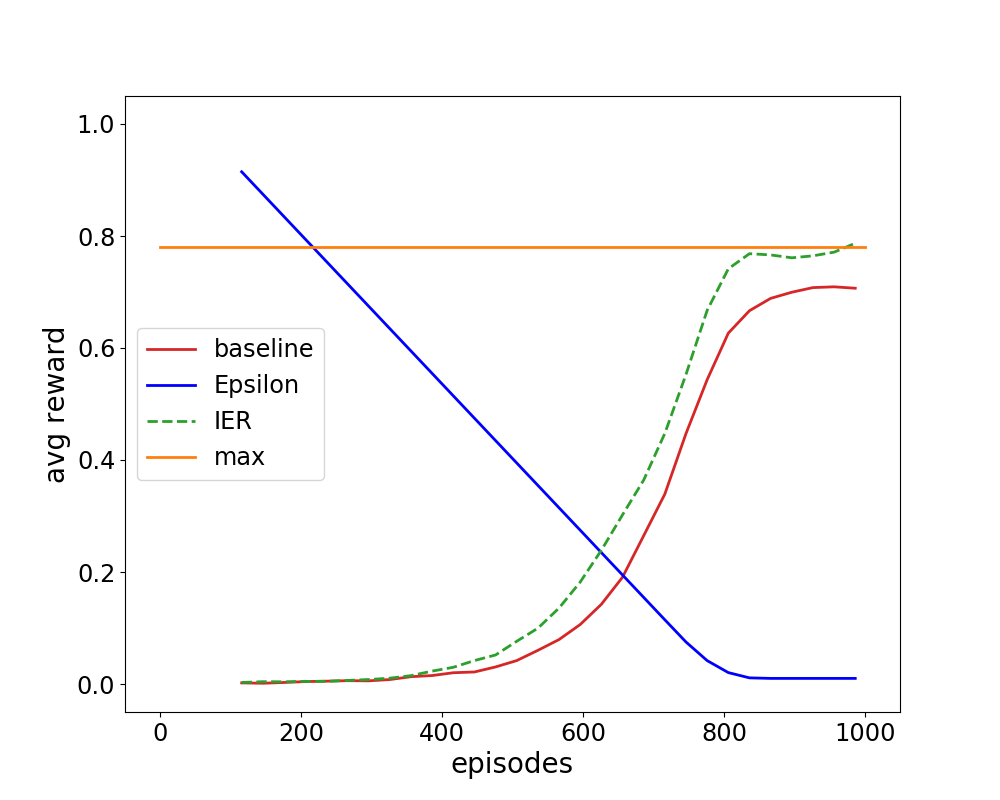}
        \caption{experiment 2}
        \label{fig:exp750}
    \end{subfigure}
    \hspace{-2.34em}
    \begin{subfigure}{.372\linewidth}
        \includegraphics[width=\textwidth]{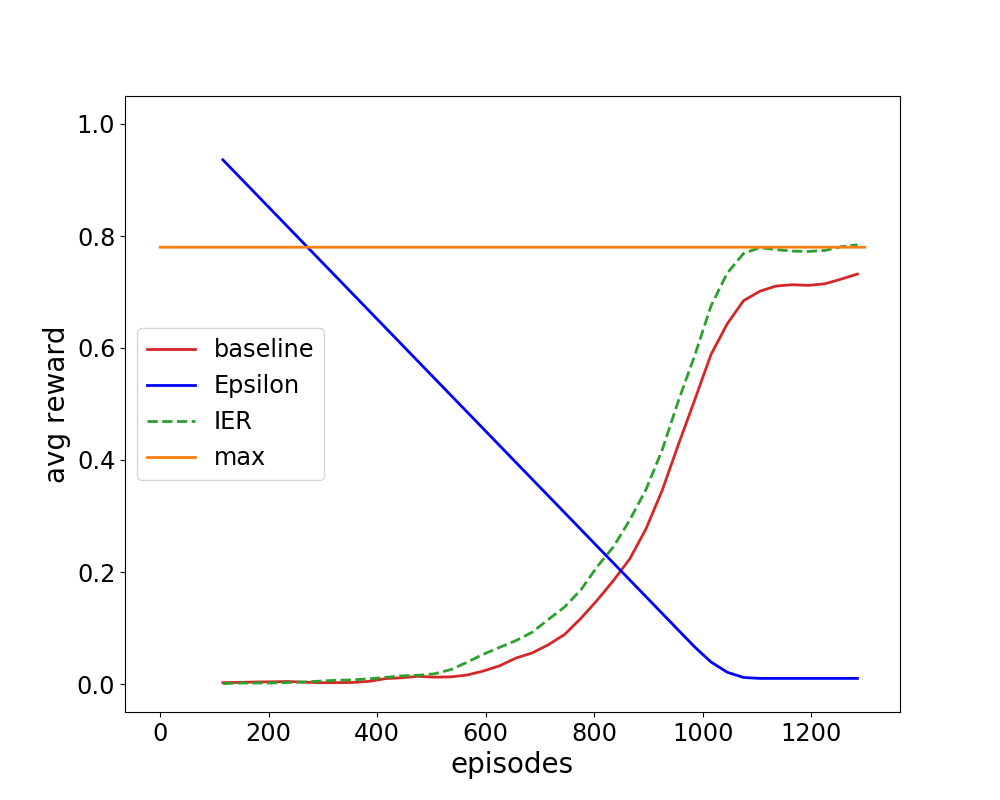}
        \caption{experiment 3}
        \label{fig:exp1k}
    \end{subfigure}
    \hspace{-2.5em}
    \caption{The best results among all condcuted experiments. The solid red line represents the classical ER serving as baseline to compare with. The dashed green line shows the average reward of the IER approach. The straight solid orange line represents the maximally reachable average reward of 0.78 and the blue line depicts the decaying epsilon. The lines for IER and the baseline represent the repetition averages. 
    }
    \label{fig:best-results}
\end{figure*}

Figure~\ref{fig:best-results} depicts the results of the best three IER configurations as given in Table~\ref{tab:best}. 
Experiment 1 and 2 were run for 1000 episodes. Experiment 3 for 1300 episodes, because of the longer exploration phase compared to the previous experiments. 
It can be observed, that the baseline approach (DQN using vanilla ER) never reaches the achievable maximum (orange line) in the observed periods and constantly stays slightly below. The IER approach, however, is converging at the maximally reachable value, and a steeper increase can be noted which indicates faster learning. 
This effect is even more distinct in the experiments with shorter exploration phases (experiments 1 and 2). 

\begin{table}[htbp]
    \centering
    \caption{Best IER configurations found during the parameter study}
    \begin{tabular}{c|c|c}
         \hline
         \textbf{experiment} & \textbf{$s_{synthetic}$} & \textbf{$c_{start\_interpolation}$}  \\
          \hline
          1 & 20,000 & 1,000 \\
          2 & 20,000 & 250 \\
          3 & 100,000 & 1,000 \\
          \hline
    \end{tabular}
    \label{tab:best}
\end{table}{}

\begin{figure*}
    \centering
    \hspace{-2.5em}
    \begin{subfigure}{.372\linewidth}
        \includegraphics[width=\textwidth]{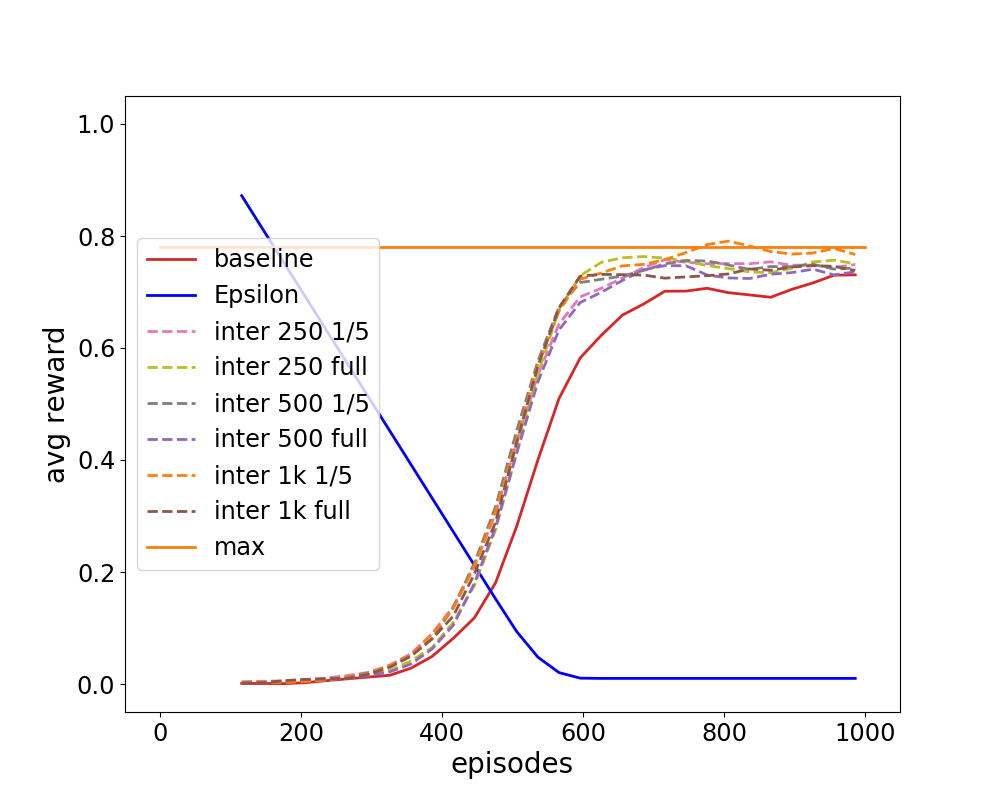}
        \caption{experiment 1}
        \label{fig:exp500-all}
    \end{subfigure}
    \hspace{-2.34em}
    \begin{subfigure}{.372\linewidth}
        \includegraphics[width=\textwidth]{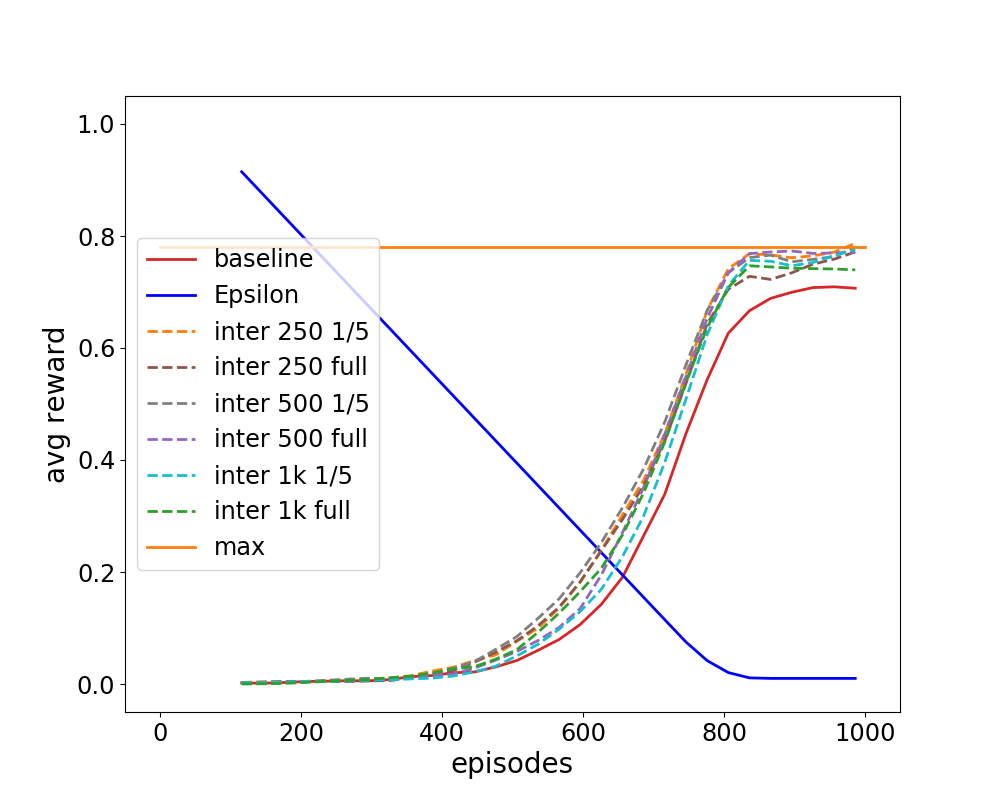}
        \caption{experiment 2}
        \label{fig:exp750-all}
    \end{subfigure}
    \hspace{-2.34em}
    \begin{subfigure}{.372\linewidth}
        \includegraphics[width=\textwidth]{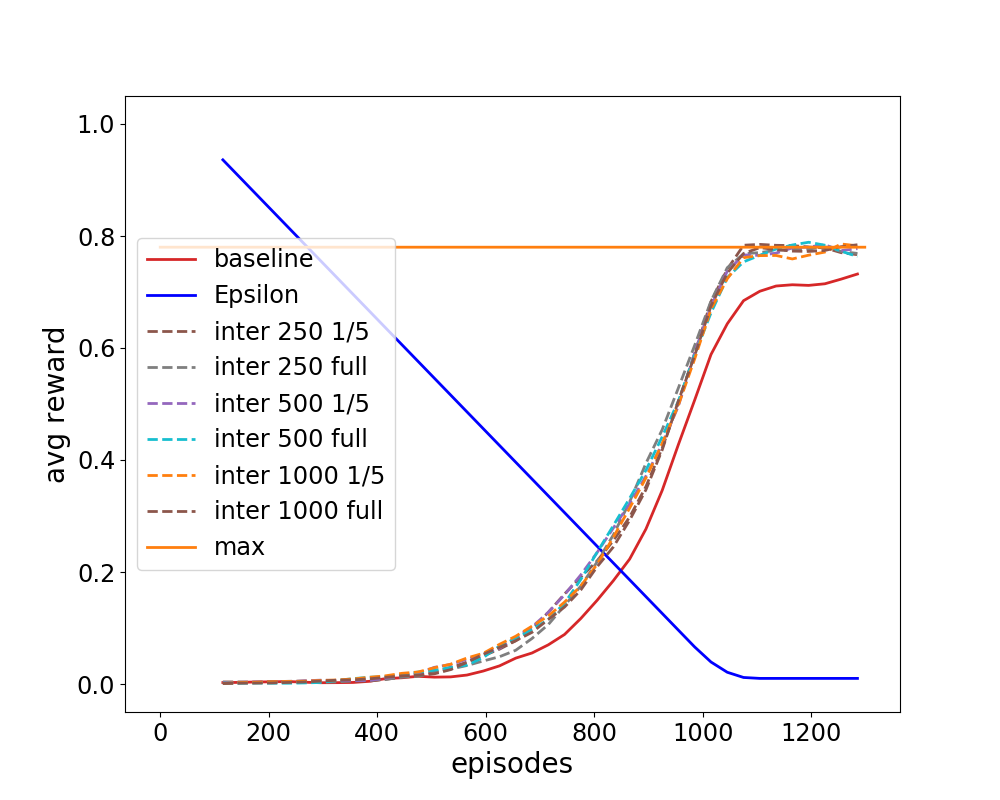}
        \caption{experiment 3}
        \label{fig:exp1k-all}
    \end{subfigure}
    \hspace{-2.5em}
    \caption{All experiments with all perturbations of $t_{exploration}$, $s_{synthetic}$ and $c_{start\_interpolation}$. The dashed lines show the results of the single experiments. The continuous straight orange line draws the maximal reachable average reward of 0.78 and the blue line depicts the decaying epsilon. The x-axis represents the episodes and the y-axis the average reward of all 20 repetitions. }
    \label{fig:all-results}
\end{figure*}

Figure~\ref{fig:all-results} reports the results of all experiments. The plots reveal, similar to Figure~\ref{fig:best-results}, that the IER approach outperforms the baseline. 
All the tested IER configurations perform similarly well, with only marginal deviations. 
It turns out that the choice of $s_{synthetic}$ and $c_{start\_interpolation}$ only has little to no effect. 
Because the IER algorithm performs better than the baseline, and this effect is even bigger in the scenarios with shorter exploration phases, it can be used to decrease the time needed for exploration, which comes in handy if exploration is costly.

\begin{figure*}
    \centering    
    \hspace{-2.5em}
    \begin{subfigure}{.372\linewidth}
        \includegraphics[width=\textwidth]{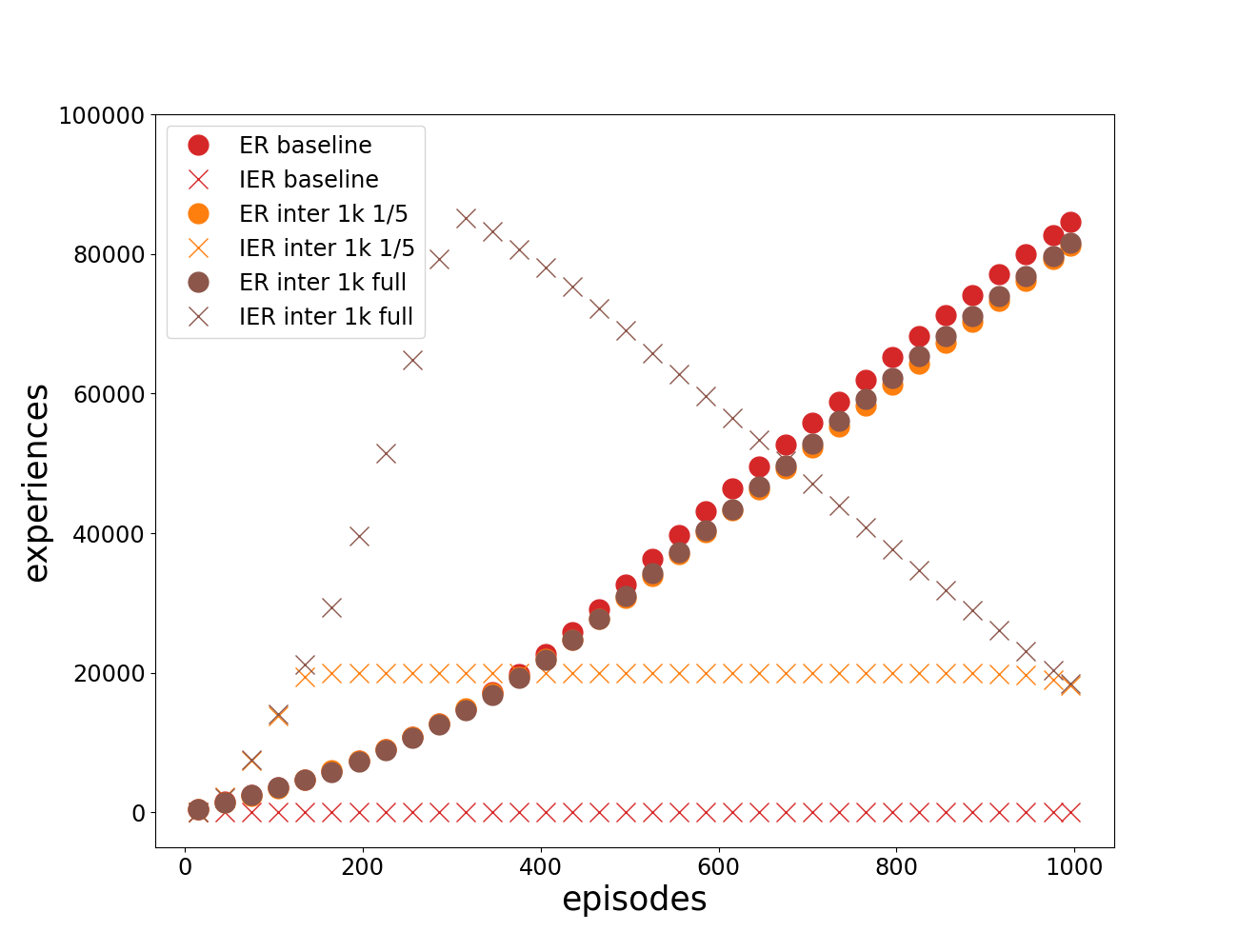}
        \caption{experiment 1}
        \label{fig:exp500-fill}
    \end{subfigure}
    \hspace{-2.34em}
    \begin{subfigure}{.372\linewidth}
        \includegraphics[width=\textwidth]{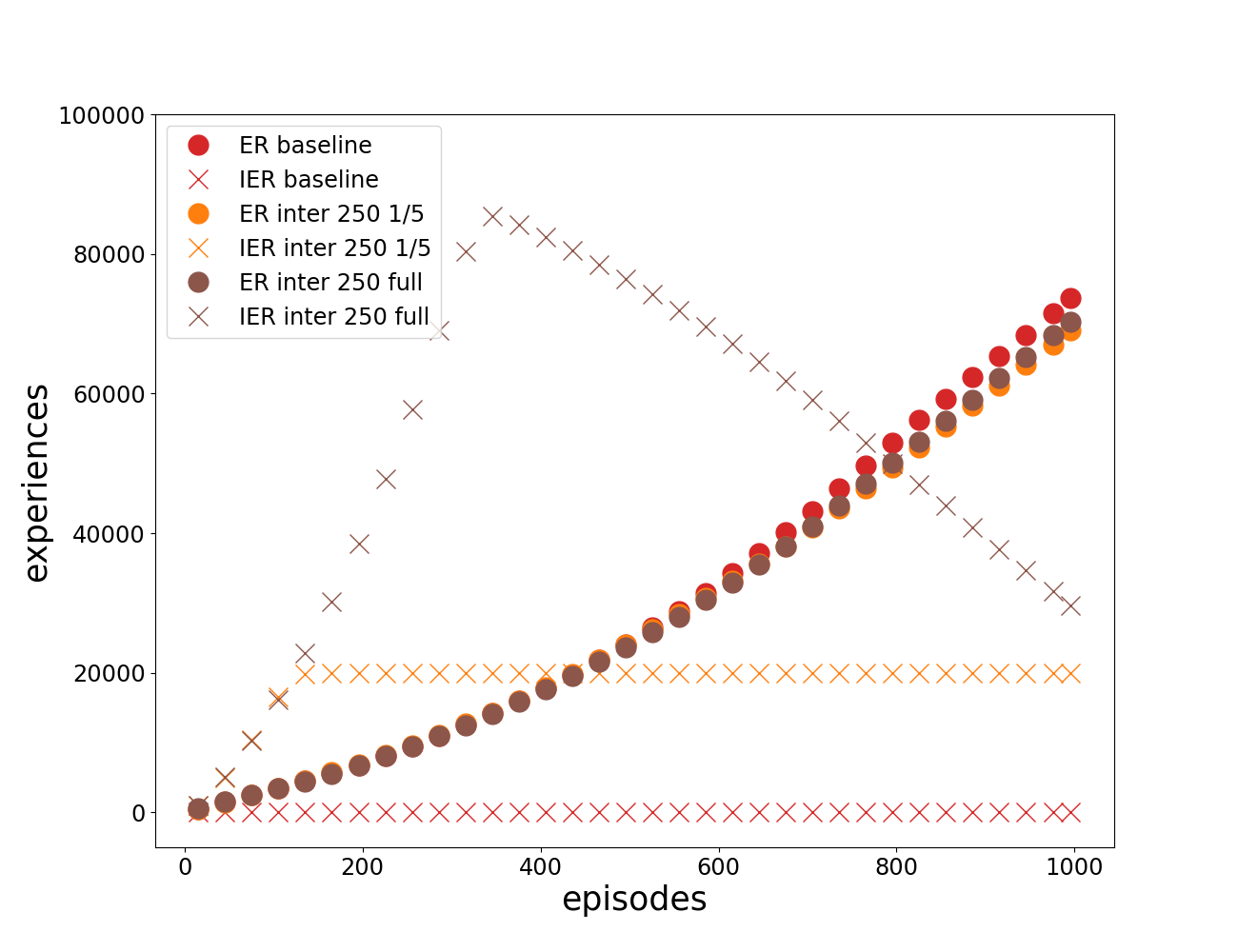}
        \caption{experiment 2}
        \label{fig:exp750-fill}
    \end{subfigure}
    \hspace{-2.34em}
    \begin{subfigure}{.372\linewidth}
        \includegraphics[width=\textwidth]{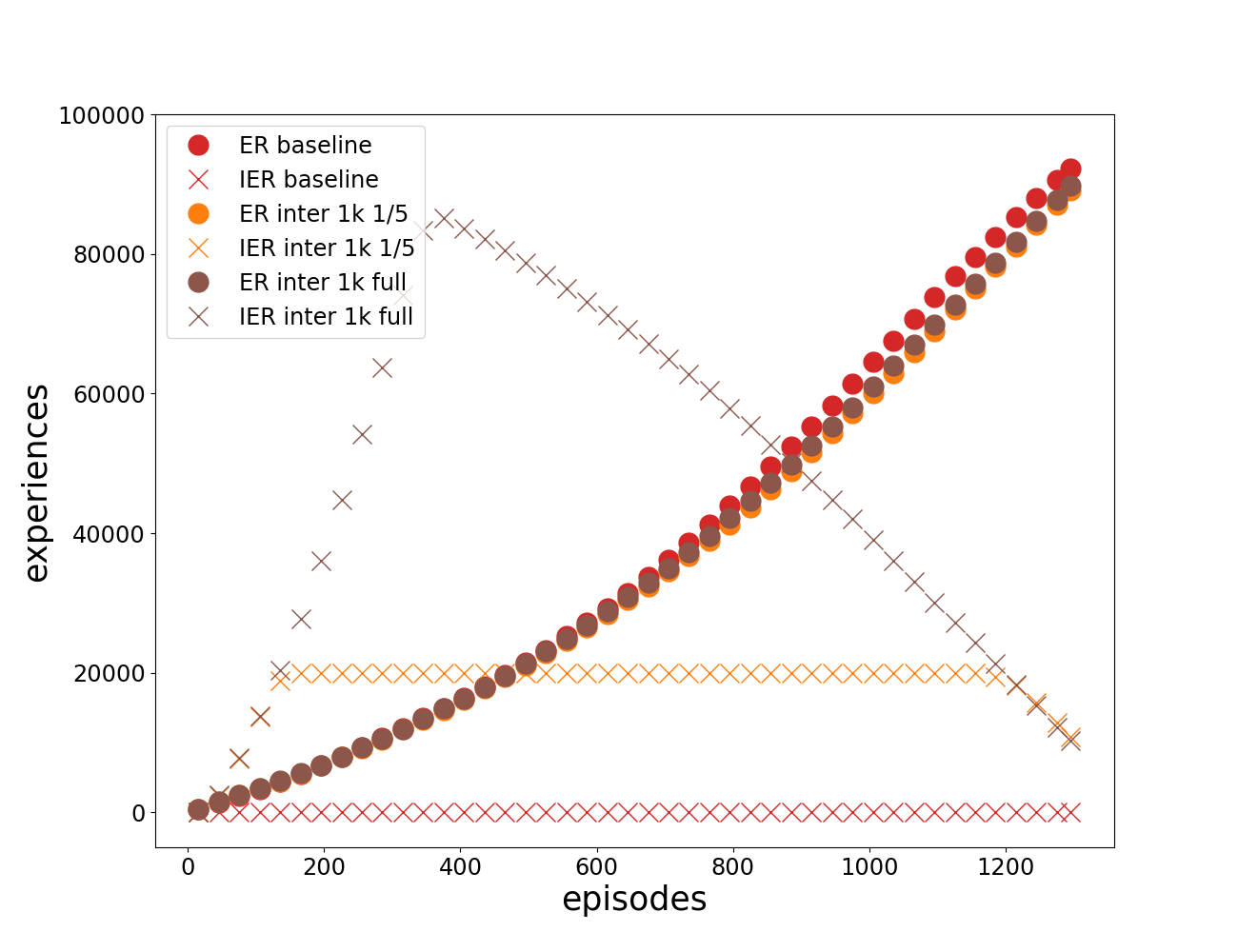}
        \caption{experiment 3}
        \label{fig:exp1k-fill}
    \end{subfigure}
    \hspace{-2.5em}
    \caption{The size of the IER represented by the amount of real and synthetic samples. $c_{start\_interpolation}$ was chosen from Tab.~\ref{tab:best}. The crosses represent the amount of synthetic and the dots the amount of real experiences. The brown curves show the size of the IER with $s_{syntehtic}=100,000$ and the orange curves with $s_{synthetic}=20,000$. The red curves represent the baseline. The x-axis marks the episode length and the y-axis the amount of stored experiences.}
    \label{fig:all-fill}
\end{figure*}

In Figure~\ref{fig:all-fill} the size of the IER can be seen. As the choice of $c_{start\_interpolation}$ does not have a huge effect on the amount of interpolated experiences compared to the maximum size we plotted only the graphs for the configurations of the best results. The crossed curves represent the amount of stored interpolated experiences and the dotted curves the amount of stored real experiences. The red curve depicts the baseline and the amount of real samples is slightly above the IER variants in all three experiments. Taking into account that, first, an episode ends after the agent has, either reached the final state, fell into a hole or reached the maximum time limit, and, second, the IER agents performed better, it seems that the baseline agent learned to avoid falling into a hole, but does not reach the final state as often as the other agents. This explains the higher amount of experiences. Fig.~\ref{fig:exp500-fill} shows that the ratio of experiences at the end of the exploration phase is in favour of the synthetic ones in the case of $s_{synthetic}=100,000$. Fig.~\ref{fig:exp750-fill} shows that the ratio changed but is still in favour of the synthetic samples and Fig.~\ref{fig:exp1k-fill} shows that at this time the ratio is in favour of the real experiences. If we look at the graphs for $s_{synthetic}=20,000$, then all ratios are in favour of the real examples, but also not that far away from a ratio of 50/50 in experiment 1 and 2. This seems to be a good choice as the best results were achieved with a choice of $s_{synthetic}$ that is close to an equal distribution of interpolated and real transitions. This should be investigated further.

\section{Conclusion and Future Work}
\label{sec:conc}

We presented an extension for the classic ER used in Deep Neural Network based RL that includes synthetic experiences to speedup and improve learning in nondeterministic and discrete environments. The proposed algorithm uses stored, actually seen transitions to utilize the experience of the model which serve as basis for the calculation of synthetic $(s,a,r,s)$ tuples by means of interpolation.  The synthetic experiences comprise a more accurate estimate of the expected long-term return a state-action pair promises, than a real sample does. So far the employed interpolation technique is a simple equally weighted averaging which serves as an initial approach. More complex methods in more complex problem spaces have to be investigated in the future. The IER approach was compared to the default ER in the FrozenLake8x8-v0 environment from the OpenAI Gym and showed an increased performance in terms of a 17\% increased overall mean reward. Several configurations for the maximum size of the stored synthetic experiences, different warm-up times for the interpolation, as well as different exploration phases were examined, but revealed no remarkable effect. Nevertheless, a ratio of 50/50 for real and synthetic samples in the IER seems promising and needs further research.

As the algorithm creates a synthetic experience for every action and every followup state there is a huge amount of samples created which could be decreased in a way that takes further knowledge into account. An example would be that only those action are considered, that the actual policy would propose in the given situation. Or only for that followup state which has the most (promising) stored samples in the storage. Also, further investigation of the composition regarding the IER seems interesting, since, as stated above, the ratio of the stored transitions might have an effect. As the evaluation was limited to the FrozenLake environments provided by OpenAI Gym, the proposed algorithm could be tested on more complex versions that differ in size and difficulty. Also a continuous version with a greatly increased state and action space is required for deeper analysis. 

The used neural network isof limited complexity and therefore bears potential as well. A network that takes only two inputs, the x- and y-coordinate, can make use of the ability to generalize over neighboring states and therefore demands for several hidden layers, rendering it a deep neural network. Another possibility of changing the network input would be to decrease the agent's knowledge from a global to a local scope. Such a network could make use of convolutional layers to learn representations in the input data. Following such an approach, the agent would only know its neighboring states and could then be evaluated on different and unknown instances of the FrozenLake environment. 

As of yet, the proposed approach is limited to discrete and nondeterministic environments. We plan to develop the IER further to solve more complex problems as well. To achieve this, a solution for the unknown followup state is needed, which could also be interpolated or even predicted by a state-transition function that is learned in parallel. Here the work from Jiang et al. \cite{8598780} could serve as a possible approach to begin with. A yet simple, but nevertheless more complex problem, because of its continuity, which is beyond the domain of grid worlds is the MountainCar problem. Other, more complex interpolation techniques have to be examined to adapt our IER approach in this environment.

\newpage

\bibliographystyle{unsrt}
\bibliography{ijcnn}

\end{document}